\documentclass[journal]{IEEEtran}

\usepackage{multirow}
\usepackage{booktabs}

\usepackage{amsmath,amsfonts}
\usepackage{algorithmic}
\usepackage{algorithm}
\usepackage{array}
\usepackage[caption=false,font=normalsize,labelfont=sf,textfont=sf]{subfig}
\usepackage{textcomp}
\usepackage{stfloats}
\usepackage{url}
\usepackage{verbatim}
\usepackage{graphicx}
\usepackage{cite}
\usepackage{hyperref}
\usepackage{color}
\usepackage{etoolbox}
\makeatletter
\patchcmd{\@makecaption}
  {\scshape}
  {}
  {}
  {}
\makeatother
\begin{document}

\title{SAMSOD: Rethinking SAM Optimization for RGB-T Salient Object Detection}

\author{Zhengyi Liu, Xinrui Wang, Xianyong Fang, Zhengzheng Tu, Linbo Wang*
\thanks{This work is supported by National Natural Science Foundation of China under Grant 62376005 (Corresponding author: Linbo Wang).}
\thanks{Zhengyi Liu, Xinrui Wang, Xianyong Fang, Zhengzheng Tu, and Linbo Wang are with Key Laboratory of Intelligent Computing and Signal Processing of Ministry of Education, School of Computer Science and Technology, Anhui University, Hefei, China(e-mail: liuzywen@ahu.edu.cn, 2325687760@qq.com, fangxianyong@ahu.edu.cn, zhengzhengahu@163.com, wanglb@ahu.edu.cn)}
}

\markboth{Journal of \LaTeX\ Class Files,~Vol.~14, No.~8, February~2025}%
{Shell \MakeLowercase{\textit{et al.}}: A Sample Article Using IEEEtran.cls for IEEE Journals}

\IEEEpubid{0000--0000/00\$00.00~\copyright~2025 IEEE}

\maketitle

\begin{abstract}
RGB-T salient object detection (SOD) aims to segment attractive objects by combining RGB and thermal infrared images. To enhance performance, the Segment Anything Model has been fine-tuned for this task.
However, the imbalance convergence of two modalities and significant gradient difference between high- and low- activations are ignored,  thereby leaving room for further performance enhancement. In this paper, we propose a model called \textit{SAMSOD}, which utilizes unimodal supervision to enhance the learning of non-dominant modality and employs gradient deconfliction to reduce the impact of conflicting gradients on model convergence. The method also leverages two decoupled adapters to separately mask high- and low-activation neurons, emphasizing foreground objects by enhancing background learning.
Fundamental experiments on RGB-T SOD benchmark datasets and generalizability experiments on scribble supervised RGB-T SOD, fully supervised RGB-D SOD datasets and full-supervised RGB-D  rail surface defect detection all demonstrate the effectiveness of our proposed method. \href{https://github.com/liuzywen/SAMSOD}{https://github.com/liuzywen/SAMSOD}

\end{abstract}


\section{Introduction}
RGB-T salient object detection (SOD) \cite{wang2021salient,tu2022rgbt} aims to  identify and segment the most attractive objects  with the help of RGB and thermal infrared images, especially in certain extreme conditions, such as nighttime, foggy and rainy environment. Thermal infrared images can provide strong contrast information on temperature changes, thus compensating for the shortcomings of RGB images due to high noise, low light, and high exposure, making RGB-T SOD a topic of practical significance.

\begin{figure}[!htp]
\center
  \includegraphics[width=1\linewidth]{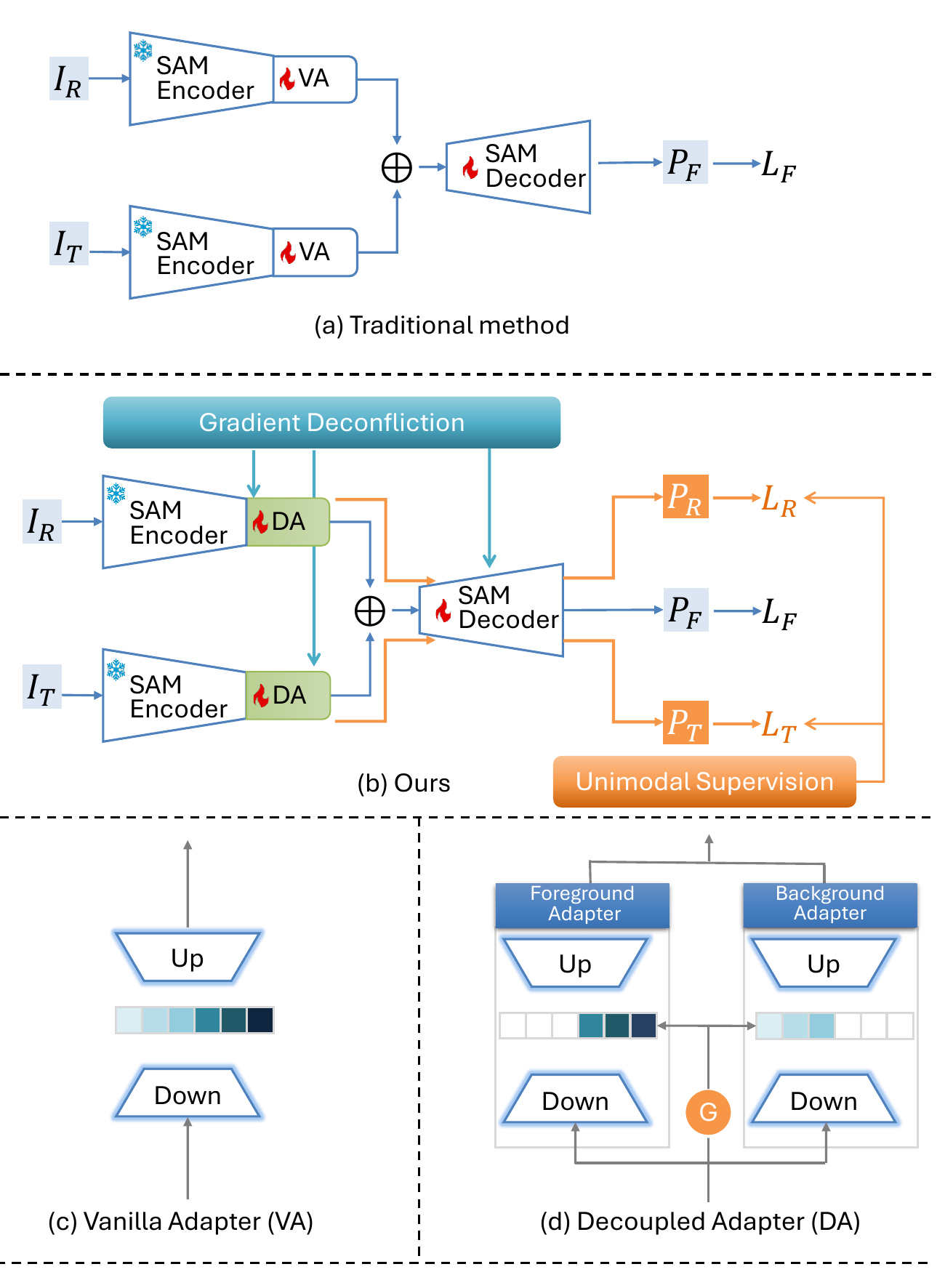}
  \caption{The framework comparison between traditional methods  and ours   and their adapters.}
  \label{fig:FrameComp}
\end{figure}

Segment Anything Model (SAM) \cite{kirillov2023segment}\cite{ravi2024sam} has emerged as a powerful segmentation model \cite{gao2024multi,zhang2024fantastic,liandiving}  by leveraging the advanced transformer architecture and being trained on a large volume of  data.
The common practice of fine-tuning SAM for RGB-T segmentation task \cite{wang2024adapting,liu2025cpal} is shown in Fig \ref{fig:FrameComp} (a). RGB and thermal images are respectively fed into two
encoders, which are a frozen and shared SAM encoder along  with two groups of finetuned adapters \cite{chen2022adaptformer}, \cite{chen2023sam},\cite{chen2024sam2}. The extracted features are fused and fed into the full-tuned SAM decoder to obtain the prediction result.

However, imbalance convergence of two modalities and significant gradient difference between high- and low- activations usually affect SAM's optimization.

Imbalance convergence of two modalities occurs in the scenario with  dominant modality \cite{peng2022balanced}\cite{weimmpareto}. Concretely, the gradient of decoder will affect  those of two encoders according to the backward propagation rule. Once an  encoder for dominant modality and the decoder converge, another encoder for non-dominant modality will be under-optimized, leading to poor multi-modal learning performance.
\IEEEpubidadjcol

To address the imbalance convergence, we introduce unimodal supervision to enhance the learning of each modality, as shown in the orange flow of Fig \ref{fig:FrameComp} (b). RGB feature, thermal feature, and fusion feature are all fed into the shared SAM decoder. Three predictions are supervised by the ground truth. Unfortunately, gradient conflict may appear in the encoders and the decoder. The gradients of unimodal supervision and fusion supervision mutually impact the RGB encoder and thermal encoder, and the ones of all supervision collaboratively influence the decoder. Once gradient conflict occurs, the training of the model may fail to converge to an ideal solution. Therefore, a gradient deconfliction strategy is applied to mitigate the impact of inconsistent gradient directions in the RGB encoder, the thermal encoder, and the decoder, respectively, as shown in the cyan flow of Fig \ref{fig:FrameComp} (b).

The significant gradient difference between high- and low- activations occurs in the vanilla adapter. Concretely, in the optimization process of SAM,  the adapter (Fig \ref{fig:FrameComp} (c)), a bottleneck structure which consists of a down-sampling, an activation, and a up-sampling,  is responsible for learning modal-specific features. The neurons after activation can be divided into two groups. High-activation neurons provide important information related with the learning object while low-activation neurons encode background cues \cite{li2024emergence}.
The derivative of the activation function for high-activation neurons is close to 1, allowing the gradient to be fully backpropagated, whereas for low-activation neurons, the derivative is close to 0, making effective parameter updates impossible. In other words, the model tends to focus on learning the foreground while ignoring the background, causing insufficient background learning. However, in many cases, learning the background can, in turn, guide the learning of the foreground. For example, when a silver object appears in a kitchen background, it is more likely to be recognized as a pot rather than a wheel.

To address significant gradient difference between high- and low- activations,   decoupled adapters are designed to synchronously strengthen the learning of foreground and background, as shown in Fig \ref{fig:FrameComp} (d).  Foreground adapter is responsible for the learning of high-activation neurons, while background adapter is in charge of the training of low-activation neurons.
By separately masking high- and low- activations, gradient interference from high-activation regions on the low-activation pathway is reduced, ensuring that low-activation features are adequately updated during training. Furthermore,
the high- and low- activation levels are determined by a learnable gate represented by the orange `G', eliminating the need to manually set the thresholds for activation.


Our contributions can be summarized as follows:
\begin{itemize}
\item We rethink SAM optimization for the RGB-T salient object detection task and propose that the issues of imbalance convergence between the two modalities and significant gradient difference between  high- and low- activations significantly affect  optimization performance.
\item To address the imbalanced convergence of the two modalities, unimodal supervision and subsequent gradient deconfliction are employed to enhance training robustness, particularly by reinforcing the learning of weaker modality and preventing the model from training instability.
\item To address significant gradient difference between high- and low- activations, decoupled adapters are utilized  to simultaneously  emphasize on the learning of foreground and background.
\item The method achieves state-of-the-art performance on three RGB-T SOD datasets, and obtains the consistent and excellent results on weakly supervised RGB-T and full supervised RGB-D SOD, and full-supervised RGB-D  rail surface defect detection, demonstrating the generalization ability.
\end{itemize}

\section{Related Works}
\subsection{RGB-T Salient Object Detection}
The salient object detection relying solely on RGB images remains susceptible to imaging conditions such as illumination variations, rainy or foggy weather, etc. Incorporating the thermal infrared sensors can provide temperature information to solve these problems to a certain extent.

Existing methods focus on model structures to achieve multi-modal fusion. MITF-Net \cite{chen2022modality} employs structural similarity multi-modal fusion, cross-level attention aggregation, and edge guidance. CGMDRNet \cite{chen2022cgmdrnet} achieves intrinsic consistency feature fusion via reducing the modality differences. CMDBIF-Net\cite{xie2023cross}  introduces an additional interactive branch and  double bidirectional interaction among three encoder branches. CAVER \cite{pang2023caver} constructs a transformer framework and aligns two modalities from spatial and channel views. HRTransNet\cite{tang2022hrtransnet} introduces HRFormer backbones and utilizes the attention mechanism to improve the performance.
PATNet \cite{jiang2024patnet} designs sophisticated  modules to emphasize on patch-level and pixel-level complementarity, pursuing the completeness and detailed refinement.
TNet \cite{cong2022does} analyzes the role of thermal modality and controls the fusion between two modalities via illuminance score.
XMSNet\cite{wu2023object} addresses sensor noise and misalignment issue by mining the cross-modal semantics.
LAFB\cite{wang2024learning} constructs an adaptive fusion bank.
Some methods, such as FFANet \cite{zhou2024frequency}, DSCDNet \cite{yu2024dual}, and WaveNet\cite{zhou2023wavenet} address multi-modal fusion from frequency perspective.
ConTriNet \cite{tang2024divide} designs a unified encoder and three specific decoders to obtain modal-specific and modal-complementary information.
SACNet \cite{wang2024alignment} proposes a new detection task for original unaligned RGB-T image pairs and gives an alignment-free solution.
SPDE \cite{jin2024underwater} tackles  salient object detection task of underwater scenes via dual-stage self-paced learning and adaptive depth emphasis.

In the paper, we propose a SAM based RGB-T model. For optimizing SAM to RGB-T model, we analyze the mechanism of multi-modal learning in encoder-decoder framework, and introduce unimodal supervision to reduce the modality dominant, and further mitigate the conflict within encoder and decoder via gradient deconfliction.
\subsection{SAM's Optimization for Multi-modal Task}
SAM has played a crucial role in the segmentation tasks of RGB images \cite{huang2024segment}, remote-sensing images \cite{yan2023ringmo}, medical images \cite{gao2024desam,shen2024fastsam3d}. Due to the scarcity of paired data, RGB-T foundation model is currently challenging. Therefore, fine-tuning SAM for RGB-T multi-modal task is a more feasible option. There are two mainstream solutions. One is auxiliary modality injection method.
SE-Adapter \cite{yao2024sam} feeds event modality into SAM via attention-like adapter.
Sammese \cite{wang2024adapting} injects the multi-modal information via a multi-modal adapter and further generates the useful prompts to help SAM segment saliency-related regions.
OpenRSS \cite{zhao2024open} finetunes SAM via thermal information prompt and dynamic low-rank adaptation in open-vocabulary RGB-T semantic segmentation task.
GoPT \cite{he2024prompting} further emphasizes on the grouping prompt tuning.
The other is two-stream parallel paradigm. SSFam \cite{liu2024ssfam} uses two finetuned SAMs to extract multi-modal features, achieving a  scribble supervised SOD method. CPAL \cite{liu2025cpal} uses the bi-directional cross-prompting tuning to optimize the foundation model.

The aforementioned methods elaborately design model structure and finetuning strategy. However, imbalance convergence optimization of two modalities is not addressed.
In the paper, we rethink SAM's optimization for two-stream segmentation model.

\subsection{Imbalance Optimization in Multi-modal Learning}
Modality dominance in multi-modal learning has received widespread attention.
One modality can end up dominating the learning process, restricting the effective use of information from other modalities and resulting in suboptimal model performance.
OGM-GE \cite{peng2022balanced} points out optimization imbalance issue in multi-modal classification task and proposes to modulate the gradient of each modality via contribution discrepancy to the learning objective.
PMR \cite{fan2023pmr} introduces prototype to address the same issue.
AGM \cite{li2023boosting} improves OGM-GE to accommodate complex fusion strategies and more modalities.
ReGrad \cite{lin2024suppress} reduces the influence of  dominant modality via modulating both conflicted and unconflicted gradients.
CGGM \cite{guoclassifier} focuses on both the magnitude and direction  of gradient.
MLA \cite{zhang2024multimodal} uses an alternating unimodal learning process to minimize interference between modalities.
GDNet \cite{wei2024gradient} decouples gradient into probabilistic distribution to alleviate the gradient coupling between modalities, promoting the optimization of each modality.
MMPareto \cite{weimmpareto} proposes the optimization in multitask-like multimodal framework considering both gradient direction and gradient magnitude via Pareto idea which finds a trade-off gradient beneficial for all objectives.
DRL \cite{wei2024diagnosing} leverages varying degrees of re-initialization on the encoders of different modalities to unlock the full potential of each modality.
Fuller \cite{huang2023fuller} calibrates gradient across tasks and modalities, respectively.
Wei et al \cite{wei2024enhancing} achieve a balanced integration of modalities by addressing the sample-level modality discrepancy.

Inspired by aforementioned methods, we analyse modality dominant issue in the two-stream encoder-decoder framework for segmentation task and give the corresponding solution.

\section{Method}
\begin{figure}[!htp]
\center
  \includegraphics[width=1\linewidth]{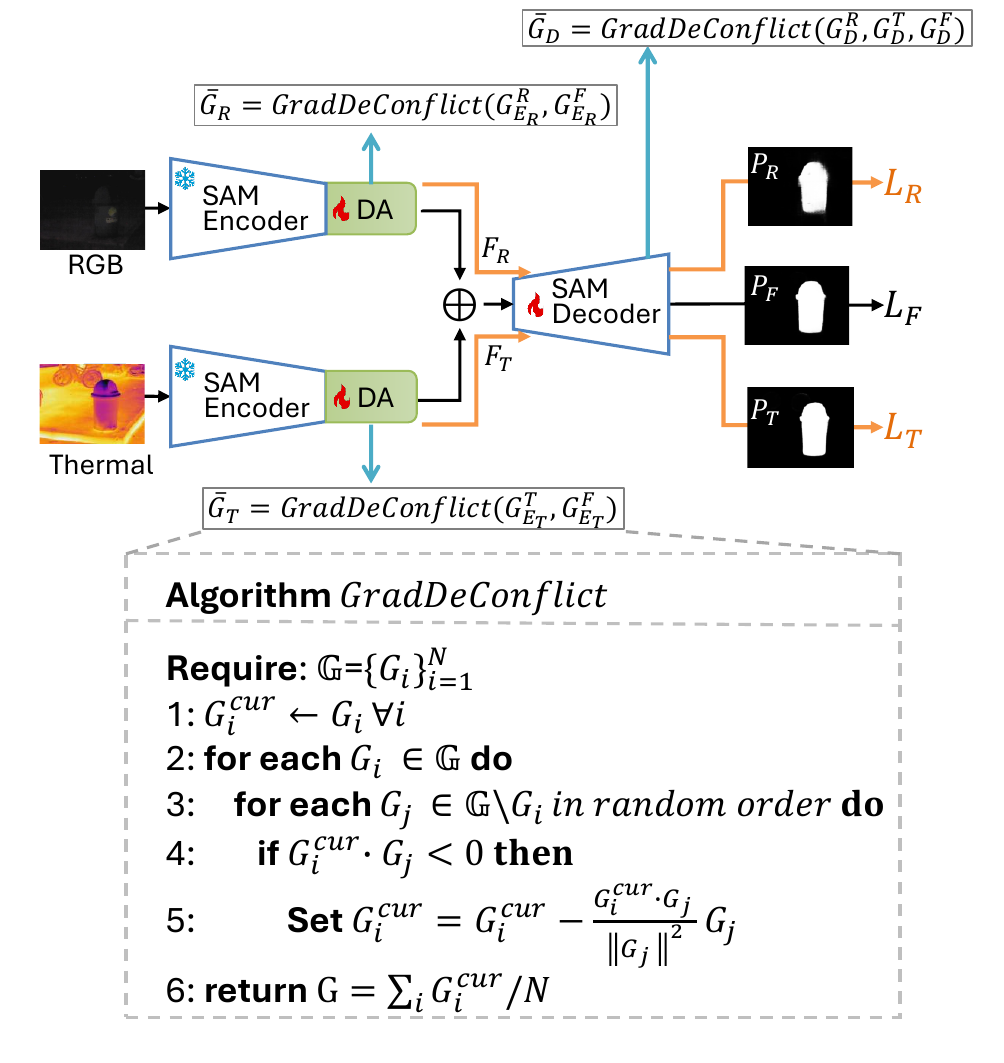}
  \caption{The pipeline of the proposed \textit{SAMSOD}.}
  \label{fig:Main}
\end{figure}

To adapt SAM for the RGB-T salient object detection task, SAM is optimized from two perspectives, resulting in a model called \textit{SAMSOD}, as shown in Fig \ref{fig:Main}. It introduces the unimodal supervision (orange flow) and gradient deconfliction (cyan flow) to address the imbalance convergence issue of RGB and thermal modalities. Moreover, it designs the decoupled adapters (green block) to solve the significant gradient difference between  high- and low- activations issue.

\subsection{Imbalance Convergence and Solution}
Existing methods address RGB-T salient object detection task via the encoder-decoder framework.
Specifically, as shown in the black flow of Fig \ref{fig:Main}, the two-stream encoder takes as input the paired RGB and thermal images $\{I_R,I_T\}$ and outputs multi-scale RGB features $F_R$ and thermal features $F_T$, respectively.
\begin{equation}
\begin{aligned}
F_R=\mathcal{E} _R (I_R;\theta_R)\\
F_T=\mathcal{E} _T (I_T;\theta_T)
\end{aligned}
\end{equation}
where $\mathcal{E} _R$ and $\mathcal{E}_T $ are the RGB encoder and thermal encoder. Here SAM2 \cite{ravi2024sam} is adopted as the encoder due to its  advanced performance in image segmentation task. Following the common practice, SAM2 encoder is frozen and shared, along with two groups of adapters with the corresponding parameters $\theta_R$ and $\theta_T$ to be finetuned.

Then the two features $F_R$ and $F_T$ are fused and fed into  the  decoder  to generate  prediction $P_{F}$.
\begin{equation}
\begin{aligned}
P_F=\mathcal{D} (F_R+F_T;\theta_\mathcal{D})
\end{aligned}
\end{equation}
where $\mathcal{D}$ is the decoder with parameter $\theta_\mathcal{D}$.

Further, the loss between the  prediction $P_F$ and ground truth $GT$ are calculated.
\begin{equation}
\begin{aligned}
\mathcal{L}_{F} = \mathcal{L}_{wiou}(P_{F}, GT)+\mathcal{L}_{wbce}(P_{F}, GT)
\end{aligned}
\end{equation}
where $\mathcal{L}_{wiou}$ and $\mathcal{L}_{wbce}$ are the weighted intersection-over-union (IoU) loss  and the weighted binary cross-entropy (BCE) loss \cite{wei2020f3net}, $\mathcal{L}_{F}$ is  called the fusion loss.

According to chain rule, the gradients of two encoders are represented as:
\begin{equation}\label{eq:chain1}
\begin{aligned}
G_{R}=\frac{\partial \mathcal{L}_F}{\partial \theta_R}=\frac{\partial \mathcal{L}_F}{\partial P_F}\cdot\frac{\partial P_F}{\partial \theta_\mathcal{D}}\cdot\frac{\partial \theta_\mathcal{D}}{\partial F_R}\cdot\frac{\partial F_R}{\partial \theta_R}\\
G_{T}=\frac{\partial \mathcal{L}_F}{\partial \theta_T}=\frac{\partial \mathcal{L}_F}{\partial P_F}\cdot\frac{\partial P_F}{\partial \theta_\mathcal{D}}\cdot\frac{\partial\theta_\mathcal{D}}{\partial F_T}\cdot\frac{\partial F_T}{\partial \theta_T}
\end{aligned}
\end{equation}

\begin{figure}[!htp]
\center
  \includegraphics[width=0.95\linewidth]{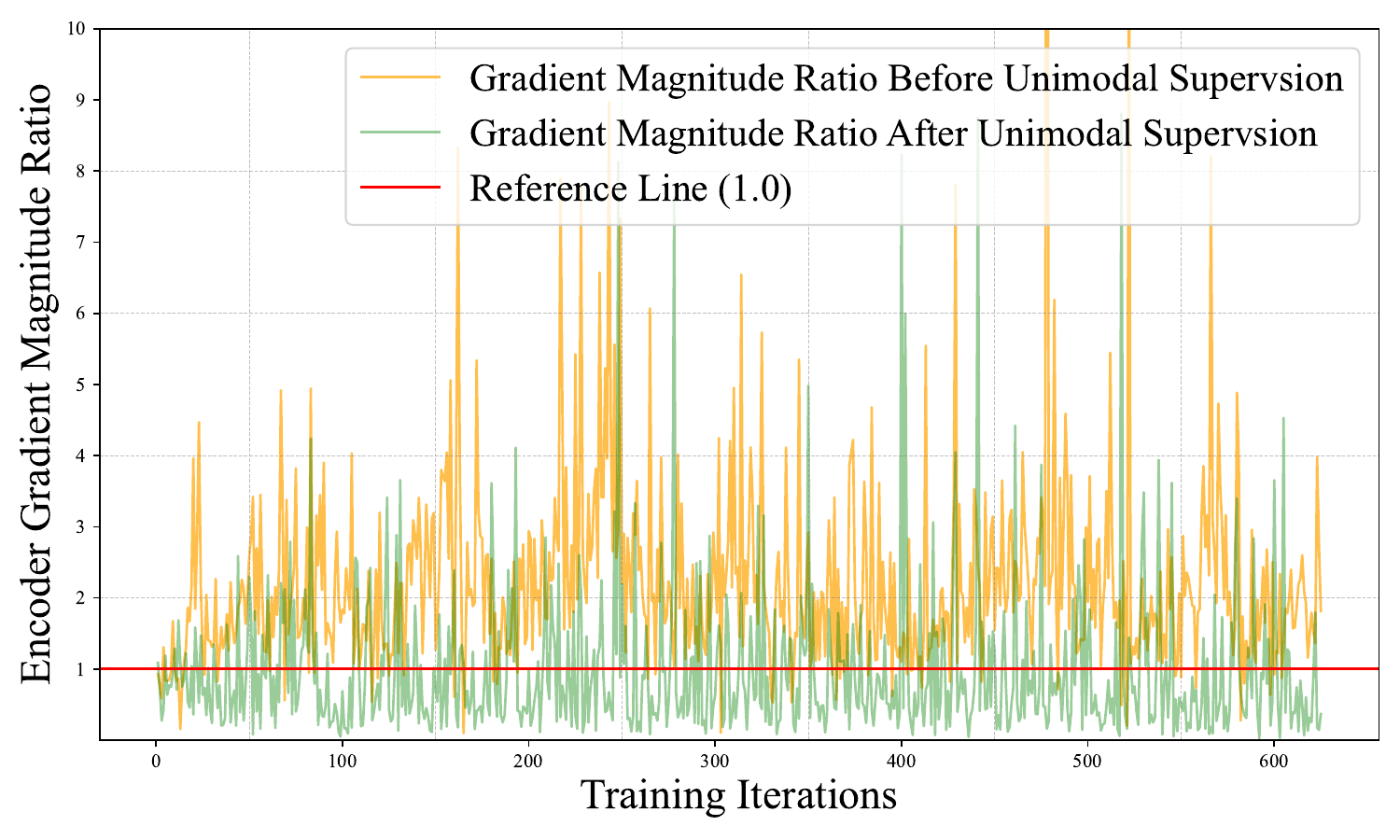}
  \caption{Gradient magnitude ratio between RGB encoder and thermal encoder.}
  \label{fig:gradRatio}
\end{figure}

Due to the differences between the RGB and thermal modalities, the last two terms of  Formula \ref{eq:chain1} are different, resulting in  $G_R\neq G_T$. The modality with the larger gradient tends to converge earlier, while the one with the smaller gradient may not reach convergence, leading to an imbalanced convergence problem. Fig \ref{fig:gradRatio} draws the gradient relation between $G_{R}$ and $G_{T}$. Obviously, the yellow curve is above the red horizontal line representing a ratio of 1, indicating $G_{R}$ is greater than $G_{T}$ in most cases.  The parameters of the RGB encoder are updated more quickly than those of the thermal encoder, as shown by the following parameter update formula:
\begin{equation}
\begin{aligned}
\theta^{t+1}_R=\theta^{t}_R-\eta*G_{R}\\
\theta^{t+1}_T=\theta^{t}_T-\eta*G_{T}
\end{aligned}
\end{equation}
where $\eta$ is learning rate, and $t$ is training iteration step.
The difference in parameter update speed indicates the RGB modal dominates the training of the whole model and the thermal modal is under-optimized so that two modalities are not well utilized, which achieves sub-optimal performance.

To address the challenge, we introduce unimodal supervision to enhance the optimization of thermal modal, as shown in the orange flow of Fig \ref{fig:Main}.
Concretely, the shared decoder is used to decode the RGB feature and thermal feature to obtain unimodal prediction $P_R$ and $P_T$.
\begin{equation}
\begin{aligned}
P_R=\mathcal{D} (F_R;\theta_\mathcal{D}), P_T=\mathcal{D}(F_T;\theta_\mathcal{D})
\end{aligned}
\end{equation}
Accordingly, two unimodal losses are calculated via:
\begin{equation}
\begin{aligned}
\mathcal{L}_{R} = \mathcal{L}_{wiou}(P_{R}, GT)+\mathcal{L}_{wbce}(P_{R}, GT) \\
\mathcal{L}_{T} = \mathcal{L}_{wiou}(P_{T}, GT)+\mathcal{L}_{wbce}(P_{T}, GT)
\end{aligned}
\end{equation}

The total loss is defined as:
\begin{equation}
\begin{aligned}
\mathcal{L}=\mathcal{L}_{F}+\mathcal{L}_{R}+\mathcal{L}_{T}
\end{aligned}
\end{equation}
From Fig \ref{fig:Main}, we can observe that the encoder part is traversed by two flows: one in black and the other in orange, while the decoder part is traversed by three flows: one in black and two in orange.
Formally, in the encoder $\mathcal{E}_R$ part, the gradient $G^R_{\mathcal{E}_R}$ from the unimodal loss $\mathcal{L}_R$ and the gradient $G^F_{\mathcal{E}_R}$ from the fusion loss $\mathcal{L}_F$ jointly influence parameter update, in the encoder $\mathcal{E}_T$ part, the gradient $G^T_{\mathcal{E}_T}$ from the unimodal loss $\mathcal{L}_T$  and the gradient $G^F_{\mathcal{E}_T}$ from the fusion loss $\mathcal{L}_F$ jointly influence parameter update, and in the decoder $\mathcal{D}$ part, the gradient $G^R_{\mathcal{D}}$ from RGB stream loss, the gradient $G^T_{\mathcal{D}}$ from thermal stream loss, and the gradient $G^F_{\mathcal{D}}$ from fusion stream jointly influence parameter update.

According to chain rule, the gradients of two encoders are updated as:
\begin{equation}\label{eq:chain2}
\begin{aligned}
G_{R}=G^F_{\mathcal{E}^R}+G^R_{\mathcal{E}^R}=\frac{\partial \mathcal{L}^F}{\partial \theta^R}+\frac{\partial \mathcal{L}^R}{\partial \theta^R}\\
G_{T}=G^F_{\mathcal{E}^T}+G^T_{\mathcal{E}^T}=\frac{\partial \mathcal{L}^F}{\partial \theta^T}+\frac{\partial \mathcal{L}^T}{\partial \theta^T}
\end{aligned}
\end{equation}
By the help of two unimodal losses, the gradients of RGB encoder and thermal encoder are influenced by both the fusion loss and their own respective losses indicated by the second term of Formula \ref{eq:chain2}. Each encoder is guided to extract its own most discriminative features, without overly relying on the fusion loss, thereby alleviating the convergence imbalance. The green curve in Fig \ref{fig:gradRatio} gives the gradient magnitude ratio between RGB and thermal encoders after unimodal supervision. The green curve is below the yellow curve and oscillate around 1, indicating the gradient magnitude of RGB encoder and the gradient magnitude of thermal encoder are mostly equal. This also reveals that their convergence imbalance has been alleviated.

\begin{figure}[!htp]
\center
  \includegraphics[width=1\linewidth]{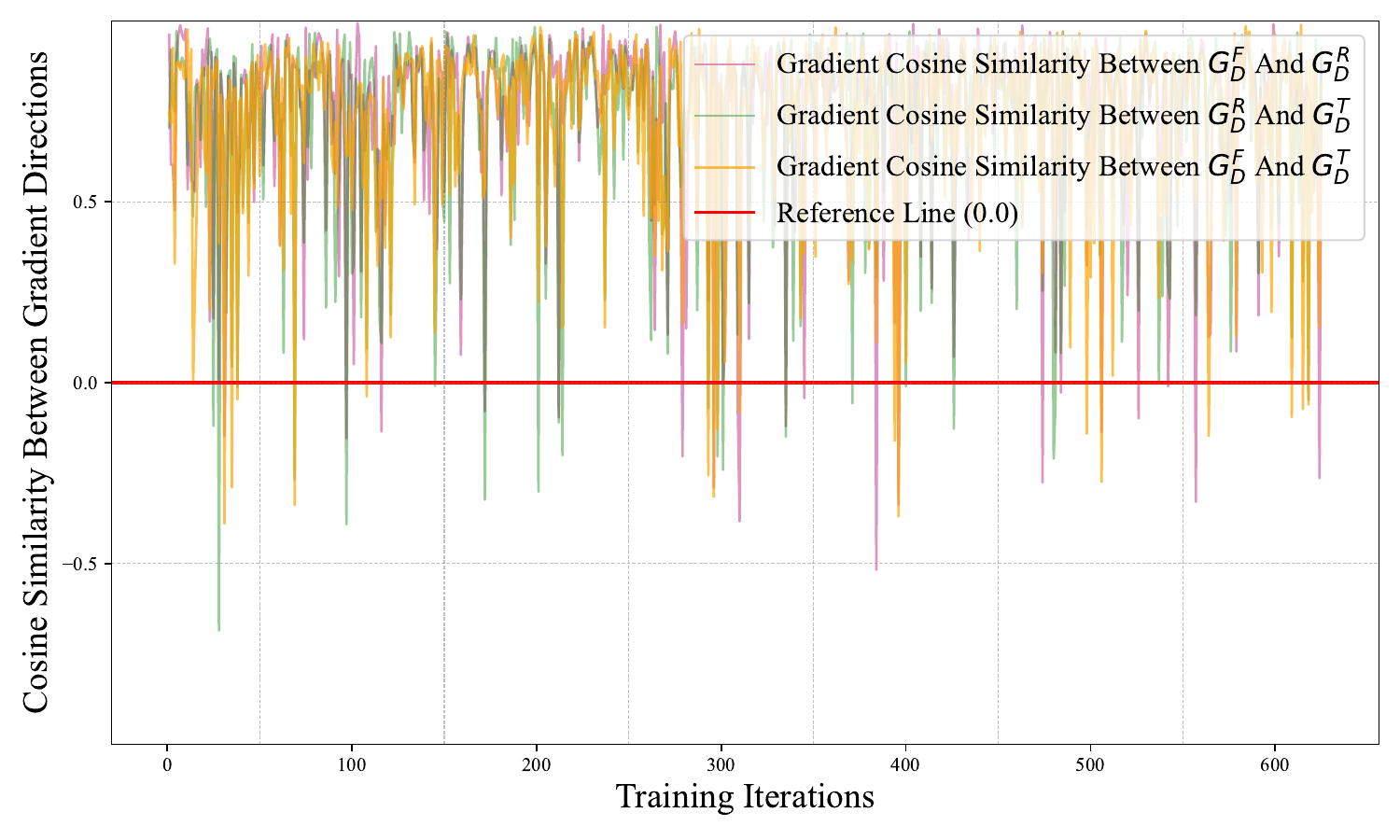}
  \caption{The  cosine similarity between gradients.}
  \label{fig:gradCosine}
\end{figure}

Similarly, the  gradient of the decoder is updated as:
\begin{equation}\label{eq:chain3}
\begin{aligned}
G_{\mathcal{D}}=G^F_{\mathcal{D}}+G^R_{\mathcal{D}}+G^T_{\mathcal{D}}=\frac{\partial \mathcal{L}^F}{\partial \theta^\mathcal{D}}+\frac{\partial \mathcal{L}^R}{\partial \theta^\mathcal{D}}+\frac{\partial \mathcal{L}^T}{\partial \theta^\mathcal{D}}
\end{aligned}
\end{equation}
The Formulas \ref{eq:chain2} and  \ref{eq:chain3} show that both $G_R$ and $G_T$ consist of two gradient terms each, while $G_{\mathcal{D}}$ consists of three gradient terms. Take $G_{\mathcal{D}}$ as an example, if the directions of $G^R_{\mathcal{D}}$, $G^T_{\mathcal{D}}$, and $G^F_{\mathcal{D}}$ are not consistent,  the overall gradient $G_{\mathcal{D}}$ is no longer the optimal optimization direction of the three gradients, but rather a compromised direction resulting from their cancellation. It is called gradient conflict. Fig \ref{fig:gradCosine} draws the cosine similarity of different gradients among $G^R_D$, $G^T_D$, and $G^F_D$. Obviously, the cosine similarities of gradients are not always larger than 0, indicating there are some opposite gradient direction, which hinders effective optimization and slows down convergence.
To this end, a gradient deconfliction is applied to alleviate the conflict of gradients to yield three updated gradients $\bar G_{R}$, $\bar G_{T}$, and $\bar G_{\mathcal{D}}$.

\begin{equation}\label{eq:GDeConflict1}
\begin{aligned}
\bar G_{R}=\textit{GradDeConflict}(G^{R}_{\mathcal{E}^R},G^{F}_{\mathcal{E}^R})
\end{aligned}
\end{equation}
\begin{equation}\label{eq:GDeConflict2}
\begin{aligned}
\bar G_{T}=\textit{GradDeConflict}(G^{T}_{\mathcal{E}^T},G^{F}_{\mathcal{E}^T})
\end{aligned}
\end{equation}
\begin{equation}\label{eq:GDeConflict3}
\begin{aligned}
\bar G_{\mathcal{D}}=\textit{GradDeConflict}(G^{R}_{\mathcal{D}} ,G^{T}_{\mathcal{D}},G^{F}_{\mathcal{D}})
\end{aligned}
\end{equation}
where $\textit{GradDeConflict}$($\mathbb{G}$) denotes a gradient deconfliction strategy on a set of gradients $\mathbb{G}$ following PCGrad \cite{yu2020gradient}, as shown in the bottom of Fig \ref{fig:Main}.
Concretely, for a given gradient $G_i\in \mathbb{G}$,  the current gradient $G_i^{cur}$  is initialized as $G_i$.
Then, the cosine similarity between current gradient $G_i^{cur}$  and another gradient $G_j\in \mathbb{G}(j\neq i)$ is estimated.
\begin{equation}
\begin{aligned}
S_{ij} = \frac{G_i^{cur} \cdot G_j}{\| G_i^{cur} \| \| G_j \|},
\end{aligned}
\end{equation}

If the cosine similarity is less than 0, it indicates a conflict in the optimization directions of the two gradients.
$G_i^{cur}$ subtracts its projection of $G^j$, otherwise, remains unchanged.

\begin{equation}
\left\{
\begin{aligned}
&G_i^{cur} = G_i^{cur} - \frac{G_i^{cur} \cdot G_j}{\| G_j \|^2} G_j,&if \;  S_{ij}<0 \\
&G_i^{cur} = G_i^{cur},&else \\
\end{aligned}
\right.
\end{equation}
The potential conflicts between $G_i^{cur}$ and all other gradients  are eliminated in the same way.

The process is repeated among all the gradients.
Finally, all the updated gradients are averaged to generate the final gradient $G$.
\begin{equation}
\begin{aligned}
G=\sum_i^{}{G_i^{cur}}/N
\end{aligned}
\end{equation}

After eliminating the gradient conflicts in all the encoders and the decoder, the parameters of two encoders and the decoder are updated as:
\begin{equation}
\begin{aligned}
\theta^{t+1}_R=\theta^{t}_R-\eta*\bar G_{R}\\
\theta^{t+1}_T=\theta^{t}_T-\eta*\bar G_{T}\\
\theta^{t+1}_\mathcal{D}=\theta^{t}_\mathcal{D}-\eta*\bar G_{\mathcal{D} }\\
\end{aligned}
\end{equation}

\subsection{Significant Gradient Difference Between High- and Low- Activations  and Solution}
\begin{figure}[!htp]
\center
  \includegraphics[width=0.95\linewidth]{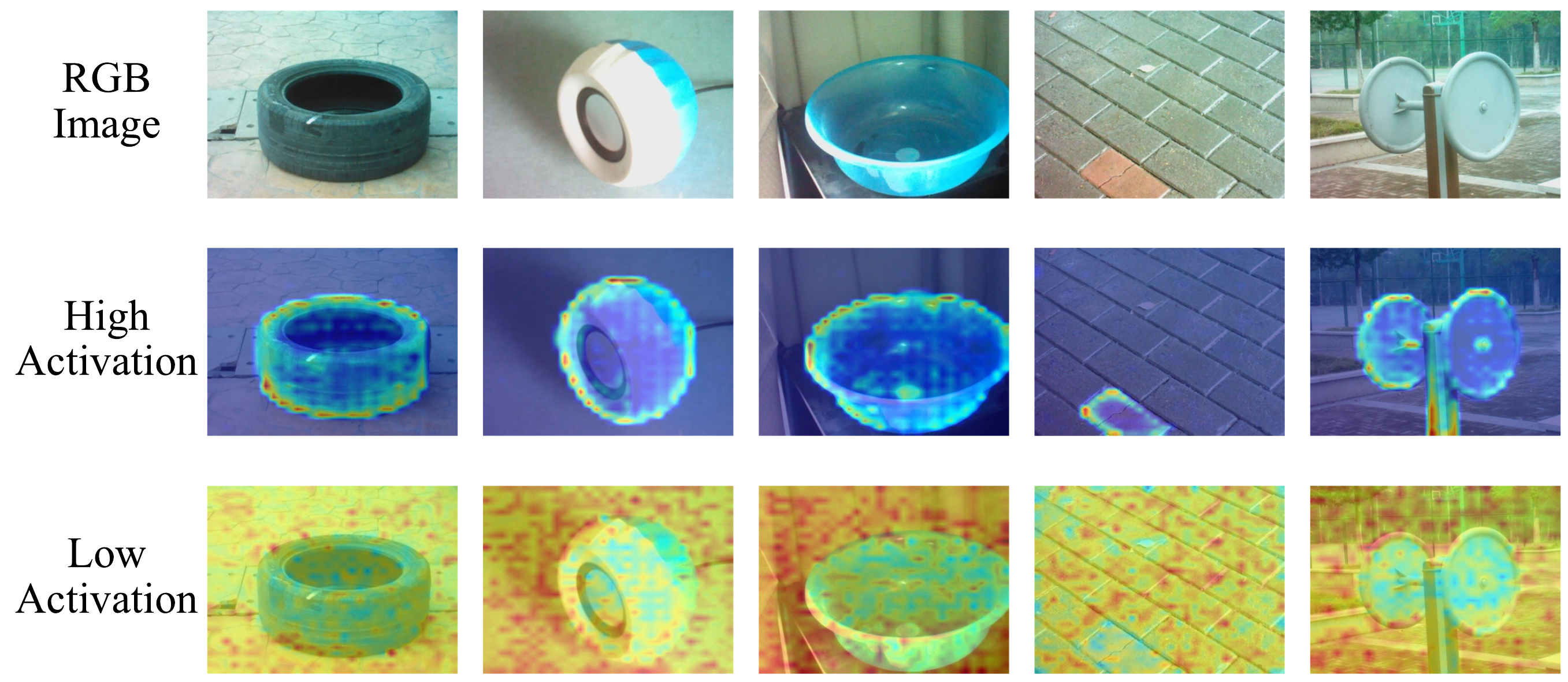}
  \caption{High- and low- activation maps of RGB images.}
  \label{fig:ActivationMap}
\end{figure}

To adapt SAM to RGB-T salient object detection task, the adapters are commonly used in practice.
Generally, for a specific input feature $x\in \mathbb{R}^{b\times n\times d}$, a  vanilla adapter is  a bottleneck structure, which includes a  down-projection layer with parameters $W_{down}\in \mathbb{R}^{d\times \hat d}$, a GeLU activation $\sigma(\cdot)$, and  an up-projection layer with parameters $W_{up}\in \mathbb{R}^{\hat d\times d}$, where $\hat d$ is the bottleneck middle dimension and satisfies $\hat d \ll d$.
\begin{equation}
\begin{aligned}
\tilde{x}= W_{up}\cdot \sigma(W_{down} \cdot x )
\end{aligned}
\end{equation}
Let $z$=$W_{down} \cdot x$ and $h$=$\sigma(z)$, the derivative of the total loss $\mathcal{L}$ with respect to $W_{down}$ is denoted as:
\begin{equation}\label{eq:dao1}
\begin{aligned}
\frac{\partial \mathcal{L}}{\partial W_{down}}=\frac{\partial L}{\partial \tilde{x}}\cdot
\frac{\partial \tilde{x}}{\partial h}\cdot \frac{\partial h}{\partial z}\cdot \frac{\partial z}{\partial W_{down}}=\frac{\partial L}{\partial \tilde{x}}\cdot W_{up}^T\cdot \sigma'(z)\cdot x^T
\end{aligned}
\end{equation}
where the first term is the  derivative of the total loss with respect to the output of  adapter $\tilde{x}$, the second term is the transpose of the up-projection weights, the third term is the derivative of the GeLU activation function on $z$, and the fourth term is the transpose of the input $x$. From the formula, we can observe that the derivative of the GeLU activation function on $z$ determines the gradient of $W_{down}$. Concretely, $\sigma'(z)$ is defined as:

\begin{equation}\label{eq:dao2}
\begin{aligned}
\sigma'(z)\approx
\begin{cases}
1, & z\gg 0\\
0.5,  & z\approx 0\\
0,& z\ll 0
\end{cases}
\end{aligned}
\end{equation}
where $\sigma'(z)\approx 1$ represents that the signal propagates freely without any suppression when $z\gg 0$, allowing the network to update the gradients normally, $\sigma'(z)\approx 0.5$ represents that the signal propagation is reduced by half when $z\approx 0$, and the network's gradient flow is partially suppressed, and $\sigma'(z)\approx 0$ represents that the signal barely passes through when $z\ll 0$, meaning the gradients of these neurons are nearly zero, making effective parameter updates impossible.
In other words, the high-activation features ($z\gg 0$) dominate gradient propagation, overshadowing low-activation information ($z\ll 0$).
Furthermore, high-activation neurons generally highlight the outline of salient objects which is important for salient object detection, while low-activation part falls more on background for the task, which is can be seen from Fig \ref{fig:ActivationMap}.
Moreover, the learning of background can provide robust context-aware cues to distinguish salient foreground objects from the background.
Therefore, to reduce gradient interference from high-activation regions on the low-activation pathway and
ensure that low-activation features are adequately updated during training, 
we decouple the adapter into a foreground adapter and a background adapter  to  synchronously enhance the learning of foreground and  background.

Concretely, the foreground adapter learns high-activation and ignores the others by dropout operation, and the background adapter learns low-activation and dropout the others. 

\begin{equation}
\begin{aligned}
\tilde{x}=W_{up}^{for}\cdot \textit{TopK}(\sigma(W_{down}^{for}\cdot  x ),P_{for},\textit{True})  \\
+ W_{up}^{back} \cdot \textit{TopK}(\sigma(  W_{down}^{back} \cdot x),P_{back},\textit{False})
\end{aligned}
\end{equation}
where $\textit{TopK}(\textit{input},k,\textit{largest})$ retains the largest  $k$ channels and dropouts the others for non-linear-transformed activated feature $\textit{input}$ when $\textit{largest}$ is $\textit{True}$, and retains the smallest  $k$ channels  and dropouts the others when  $\textit{largest}$ is $\textit{False}$, and $P_{for}$ and $P_{back}$ are used to determine the extent of foreground and background activation levels, $W_{down}^{for}$ and $W_{up}^{for}$ are parameters of foreground adapter, while $W_{down}^{back}$ and $W_{up}^{back}$ are parameters of background adapter.

In the foreground adapter, low-activation neurons are set as 0, indicating the  difference of $\sigma'(z)$ between high-activation and low-activation is  in the range of $[0.5,1]$ according to the Formula \ref{eq:dao2}. Similarly, in the background adapter, high-activation neurons are set as 0, the  difference of $\sigma'(z)$ is in the range of $[0,0.5]$. Compared with a vanilla adapter whose difference is in the range of $[0,1]$, the two decoupled adapters reduce gradient differences, preventing high-activation neurons from suppressing low-activation ones. This allows both high- and low-activation neurons to be sufficiently learned.



To adaptively determine the extent of foreground and background activation levels, $P_{for}$ and $P_{back}$ are defined as:
\begin{equation}
\begin{aligned}
P_{for} = \left\lfloor \hat d \times(\alpha_{for} + \beta \cdot G_{for}) \right\rfloor\\
P_{back} = \left\lfloor \hat d \times(\alpha_{back} + \beta \cdot G_{back}) \right\rfloor
\end{aligned}
\end{equation}
where $\left\lfloor\cdot\right\rfloor$ is floor function, $\hat d$ is bottleneck dimension with 32 channels, $\alpha_{for}$ and $\alpha_{back}$ are predefined activation ratio for the foreground adapter and background adapter, \( \beta \) is a temperature scale, and $G_{for}$ and $G_{back}$ are learnable activation ratios of foreground and background  from a  gate. Specifically,   a linear layer with weight matrix \( A\in \mathbb{R}^{n \times d} \) and bias  \( b \in \mathbb{R}^n \) and a softmax function are used to calculate the weights of the gate.
\begin{equation}
\begin{aligned}
\left[G_{for},G_{back}\right]= \textit{Softmax}(Ax+b)
\end{aligned}
\end{equation}

By separately masking high- and low- activations, this approach  reduces gradient disparities and avoid the phenomenon where strong activations dominate while weak activations are ignored.

\section{Experiments}
\subsection{Dataset and Evaluation Metrics}
Three RGB-T salient object detection datasets are used to assess the performance of the proposed method. The VT821 dataset \cite{wang2018rgb} comprises 821 manually aligned image pairs. The VT1000 dataset \cite{tu2019rgb} encompasses 1,000 RGB-T image pairs captured using highly synchronized RGB and thermal cameras. The VT5000 dataset \cite{tu2022rgbt} includes 5,000 pairs of high-resolution, diverse, and low-deviation RGB-T images. To ensure a fair comparison, we employ the same training dataset according to the usual settings, which consists of 2,500 image pairs selected from the VT5000 dataset. The remaining image pairs are reserved for testing purposes.

Four evaluation  metrics are utilized, including S-measure (S) \cite{fan2017structure}, max F-measure ($F_{\beta}$) \cite{achanta2009frequency}, max E-measure ($E_{\xi}$) \cite{fan2018enhanced}, and mean absolute error (M) \cite{perazzi2012saliency}.
\subsection{Implementation Details}
Our method is implemented based on PyTorch on a PC with an NVIDIA RTX 3090 GPU. The input image size is 512$\times$512. During the training process, photometric data augmentation is utilized for the RGB-T input data. We leverage the Adam optimizer to train our model. The max training epoch is set to 45 and the learning rate is 1e-3. Based on the large and tiny version of SAM2, two versions of the model are implemented: \textit{SAMSOD}-large and \textit{SAMSOD}-tiny. For \textit{SAMSOD}-large, the batch size is set to 4, and the training process takes approximately 15 hours. In contrast, \textit{SAMSOD}-tiny uses a batch size of 8 and requires around 4 hours for training. 
Empirically, the initial threshold $\alpha_{for}$ of foreground adapter is 0.45, the initial threshold $\alpha_{back}$ of background adapter is 0.35. The temperature scale $\beta$ is 0.1. For testing, the sum of three parallel saliency predictions is the final prediction.
\subsection{Comparison Experiments}
\subsubsection{\textbf{Comparison Methods}}
We compare our method with some RGB-T SOD algorithms, including TNet\cite{cong2022does}, HRTransNet\cite{tang2022hrtransnet}, 
CAVER \cite{pang2023caver}, WaveNet\cite{zhou2023wavenet}, PRLNet \cite{zhou2023position}, PATNet\cite{jiang2024patnet},
MAGNet \cite{zhong2024magnet}, FFANet\cite{zhou2024frequency}, ConTriNet\cite{tang2024divide}, LAFB\cite{wang2024learning}, ISMNet\cite{wang2024intra}, DSCDNet\cite{yu2024dual}, TCINet\cite{lv2024transformer}, UniTR\cite{guo2024unitr}, and SACNet\cite{wang2024alignment}. For a fair comparison, we either use the performance results reported in the original papers or evaluate the saliency maps provided by the authors using the same evaluation tools.
\begin{table*}[!htp]
\caption{Quantitative comparisons on three RGB-T SOD datasets and overall performance. `-' indicates that the corresponding cost is not reported in the original paper. The best results are bold.}
\centering

\resizebox{1\linewidth}{!}
{
   \begin{tabular}{ccccccccccccccccccccc}

\hline\toprule
   \multirow{2}{*}{\centering Methods}&\multirow{2}{*}{\centering Source}  &\multicolumn{4}{c}{\centering Overall } &\multicolumn{4}{c}{\centering VT821 } & \multicolumn{4}{c}{\centering VT1000 } & \multicolumn{4}{c}{\centering VT5000}&\multicolumn{3}{c}{\centering Cost }\\
          \cmidrule(r){3-6} \cmidrule(r){7-10} \cmidrule(r){11-14}\cmidrule(r){15-18}\cmidrule(r){19-21}
     &
     & S$\uparrow$ & F$_\beta$ $\uparrow$ &$E_{\xi}\uparrow$ & M$\downarrow$
     & S$\uparrow$ & F$_\beta$ $\uparrow$ &$E_{\xi}\uparrow$ & M$\downarrow$
     & S$\uparrow$ & F$_\beta$ $\uparrow$ &$E_{\xi}\uparrow$ & M$\downarrow$
     & S$\uparrow$ & F$_\beta$$\uparrow$&$E_{\xi}\uparrow$ & M$\downarrow$
     &Params (M)$\downarrow$&FLOPs (G)$\downarrow$&FPS (Hz)$\uparrow$\\
    \midrule

    TNet\cite{cong2022does}&TMM22
    &.904&.894&.944&.030
    &.899&.888&.938&.030
    &.929&.930&.966&.021
    &.895&.881&.937&.033
    &87.0&39.7&49.6\\
    HRTransNet \cite{tang2022hrtransnet}&TCSVT22
    &.917&.909&.958&.023
    &.906&.888&.941&.026
    &.938&.941&.975&.017
    &.912&.903&.956&.025
    &58.9&\textbf{17.3}&14.9\\
%

    CAVER \cite{pang2023caver}&TIP23
    &.908&.894&.949&.025
    &.898&.877&.934&.027
    &.938&.939&.973&.017
    &.899&.882&.944&.028
    &93.8&31.6&28.5\\
    WaveNet\cite{zhou2023wavenet}&TIP23
    &.919&.907&.955&.023
    &.912&.895&.943&.024
    &.945&.945&.977&.015
    &.911&.896&.950&.026
    &\textbf{30.2}&26.7&7.7\\
    PRLNet \cite{zhou2023position}&TIP23
    &.926&.878&.946&.022
    &.917&.860&.932&.025
    &.944&.902&.951&.016
    &.921&.875&.948&.023
    &-&-&-\\
PATNet\cite{jiang2024patnet}&KBS24
&.921&.924&.952&.021
&.914&.914&.938&.024
&.941&.948&.964&.015
&.916&.917&.951&.023
&95.9&51.1&-\\
MAGNet \cite{zhong2024magnet}&KBS24
&.916&.882&.942&.023
&.909&.865&.930&.026
&.938&.913&.949&.016
&.909&.875&.943&.025
&-&-&-\\
FFANet\cite{zhou2024frequency}&PR24
&.921&.888&.948&.021
&.905&.855&.926&.027
&.943&.918&.955&.014
&.918&.886&.953&.021
&364.3&206.5&-\\
ConTriNet\cite{tang2024divide}&TPAMI24
&.926&.899&.952&.019
&.915&.878&.940&.022
&.941&.918&.954&.015
&.923&.898&.956&.020
&96.3&126.9&-\\
LAFB\cite{wang2024learning}&TCSVT24
&.910&.899&.950&.025
&.900&.884&.940&.028
&.932&.930&.969&.018
&.904&.891&.945&.027
&453.0&139.7&45.0\\
ISMNet\cite{wang2024intra}&TCSVT24
&.920&.894&.947&.022
&.917&.886&.945&\textbf{.021}
&.942&.922&.954&.014
&.913&.885&.945&.025
&114.3&100.4&6.2\\

DSCDNet\cite{yu2024dual}&TCE24
&.924&.893&.949&.021
&.915&.876&.940&.022
&.946&.921&.955&.014
&.918&.888&.949&.023
&92.3&134.1&-\\
TCINet\cite{lv2024transformer}&TCE24
&.926&.910&.965&.018
&.914&.886&.951&\textbf{.021}
&.942&.928&.976&.014
&.924&.910&.965&.019
&88.2&91.9&25.9\\

    UniTR\cite{guo2024unitr}&TMM24
    &.913&.914&.958&.021
    &.901&.898&.941&.025
    &.938&.939&.975&.014
    &.907&.910&.956&.023
    &72.0&-&\textbf{133.3}\\
    SACNet\cite{wang2024alignment}&TMM24
    &.921&.892&.952&.020
    &.906&.859&.932&.025
    &.942&.927&.958&.014
    &.917&.888&.957&.021
    &327.7&-&27.0\\
    \textbf{\textit{SAMSOD}-tiny}&-
    &.925&.921&.964&.020
    &.913&.895&.950&.022
    &.942&.942&.976&.015
    &.923&.921&.964&.021
    &32.7&60.4&23.1\\
    \textbf{\textit{SAMSOD}-large}&-
    &\textbf{.935}&\textbf{.933}&\textbf{.969}&\textbf{.017}
    &\textbf{.924}&\textbf{.916}&\textbf{.956}&\textbf{.021}
    &\textbf{.948}&\textbf{.949}&\textbf{.981}&\textbf{.012}
    &\textbf{.934}&\textbf{.933}&\textbf{.969}&\textbf{.017}
    &224.1&418.0&6.4\\
   \bottomrule
    \hline
\end{tabular}
}
\label{tab:RGBTCom}
\end{table*}
\subsubsection{\textbf{Quantitative Analysis}}
\begin{figure}[!htp]
\begin{tabular}{ccc}
\includegraphics[width = 0.3\linewidth]{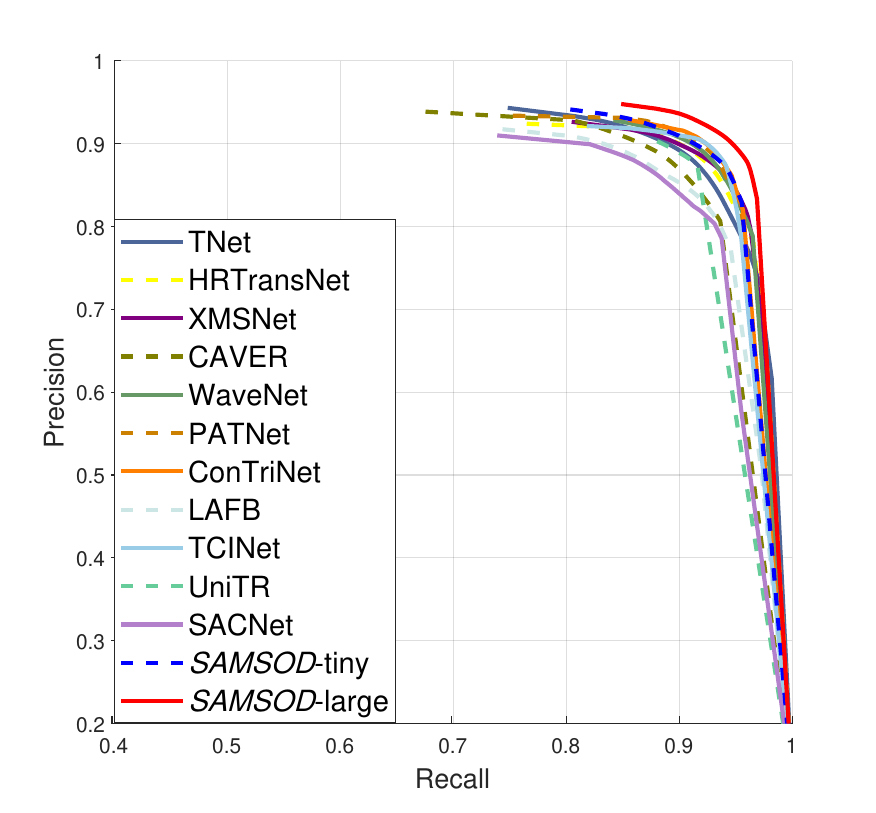}&\includegraphics[width = 0.3\linewidth]{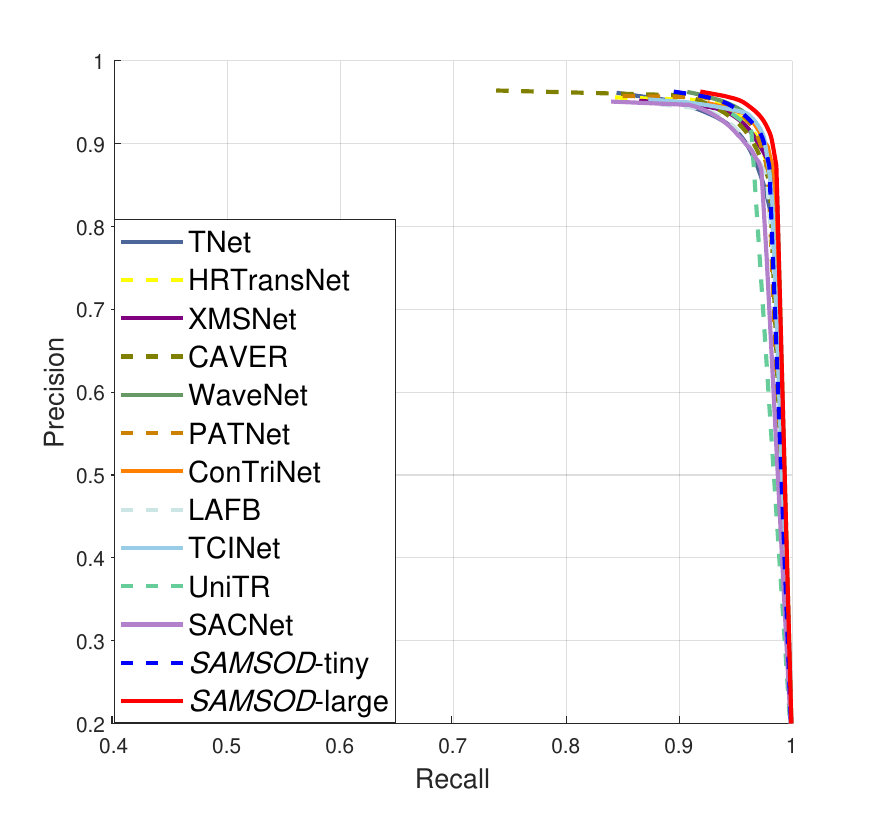}&\includegraphics[width = 0.3\linewidth]{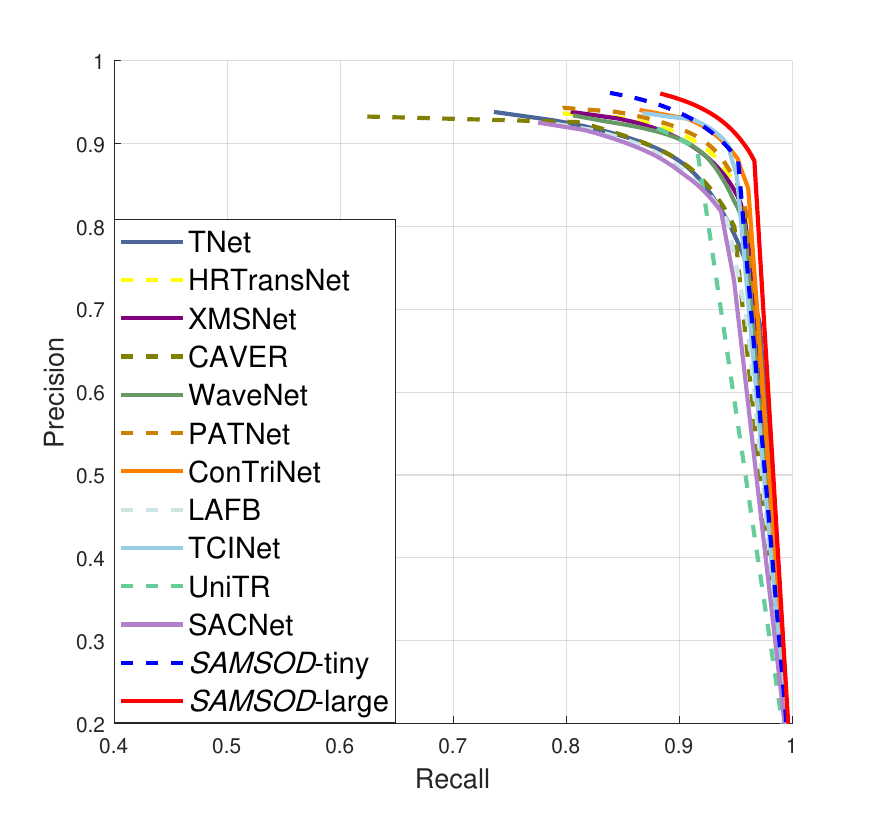}\\
\scriptsize (a) VT821 dataset&\scriptsize (b) VT1000 dataset&\scriptsize (c) VT5000 dataset\\
\end{tabular}
\caption{P-R curves comparison of different models on three RGB-T datasets. }
\label{fig:PR}
\end{figure}

From Table \ref{tab:RGBTCom}, it is clear that \textit{SAMSOD}-large achieves the best results regardless of individual dataset or overall performance.
In fact, by introducing unimodal supervision and further employing gradient deconfliction strategy, the two modalities are well utilized to jointly achieve the better segmentation performance. Meanwhile, the proposed decoupled adapters  finetuned on the frozen SAM2 encoder enhances the understanding of foreground objects by studying the background, leading to a more accurate segmentation of the foreground region.
In addition, the comparison of PR curves in Fig \ref{fig:PR} further validates the superiority of \textit{SAMSOD}-large, because our PR curve is  closer to the top-right corner indicating that both precision and recall are highest.

\subsubsection{\textbf{Qualitative Analysis}}
\begin{figure}[!htp]
\center
  \includegraphics[width=1\linewidth]{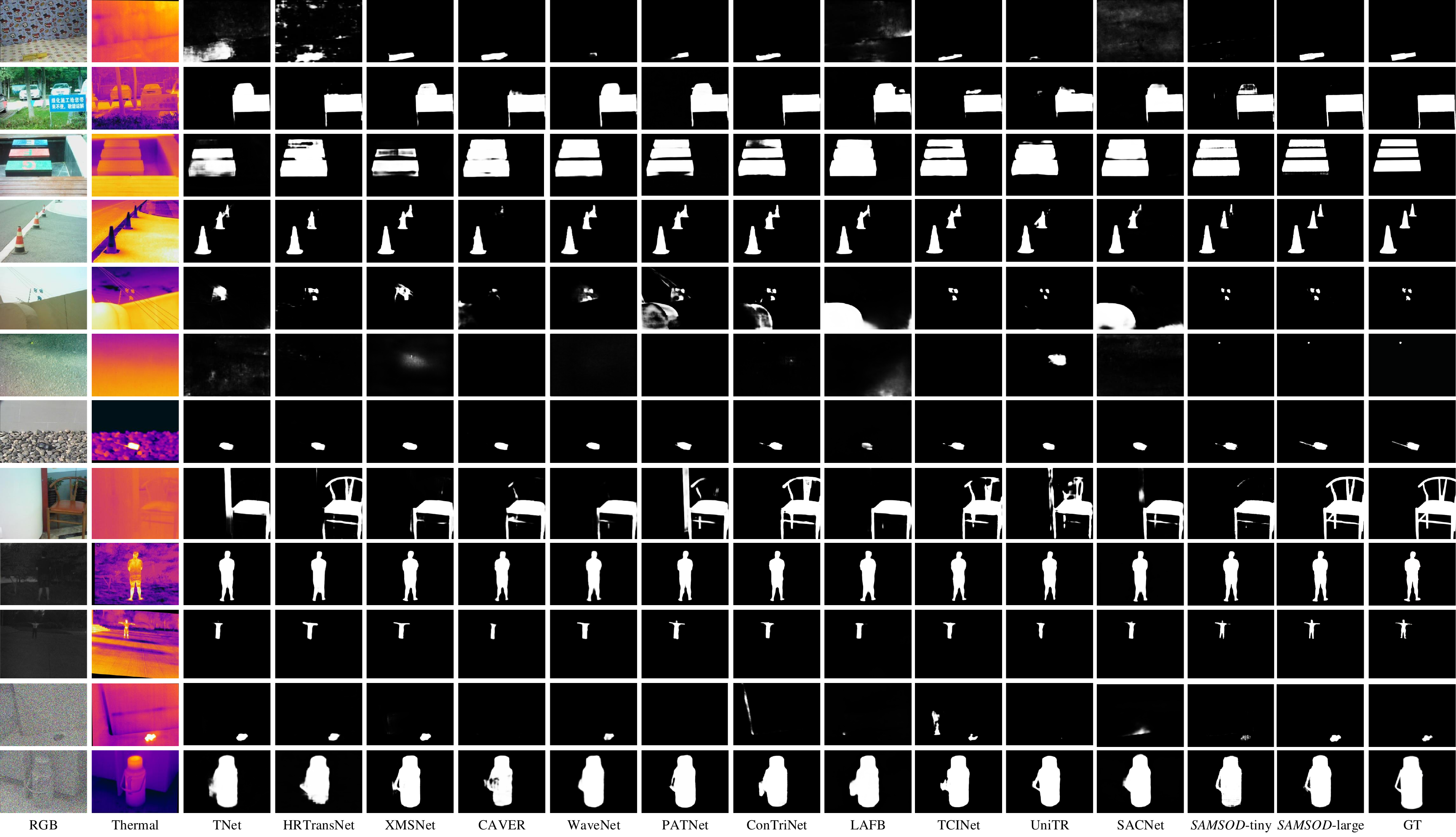}
  \caption{Qualitative comparisons of visualization in complex background ($1^{st}$ and $2^{nd}$ rows), multiple objects ($3^{rd}$ and $4^{th}$ rows),  small objects ($5^{th}$ and $6^{th}$ rows), fine-grained objects ($7^{th}$ and $8^{th}$ rows), objects in low light environment ($9^{th}$ and $10^{th}$ rows), and objects in strong noise condition ($11^{th}$ and $12^{th}$ rows)}
  \label{fig:Visual}
\end{figure}

Fig \ref{fig:Visual} presents the visualization comparison among saliency maps of \textit{SAMSOD}-large and competitors in various complex scenarios, including complex background ($1^{st}$ and $2^{nd}$ rows), multiple objects ($3^{rd}$ and $4^{th}$ rows),  small objects ($5^{th}$ and $6^{th}$ rows), fine-grained objects ($7^{th}$ and $8^{th}$ rows), objects in low light environment ($9^{th}$ and $10^{th}$ rows), and objects in strong noise condition ($11^{th}$ and $12^{th}$ rows). From the results we can observe that our method produces saliency maps that most closely resemble the ground truth across aforementioned scenes. For example, in the first two rows, foreground objects are well recognized by learning the background. In the last four rows, although the RGB modality provides limited information in low-light and high-noise environments, thermal images remain effectively utilized, demonstrating the effectiveness of unimodal supervision.

\subsubsection{\textbf{Failure Cases Analysis}}
\begin{figure}[!htp]

\center
  \includegraphics[width=1\linewidth]{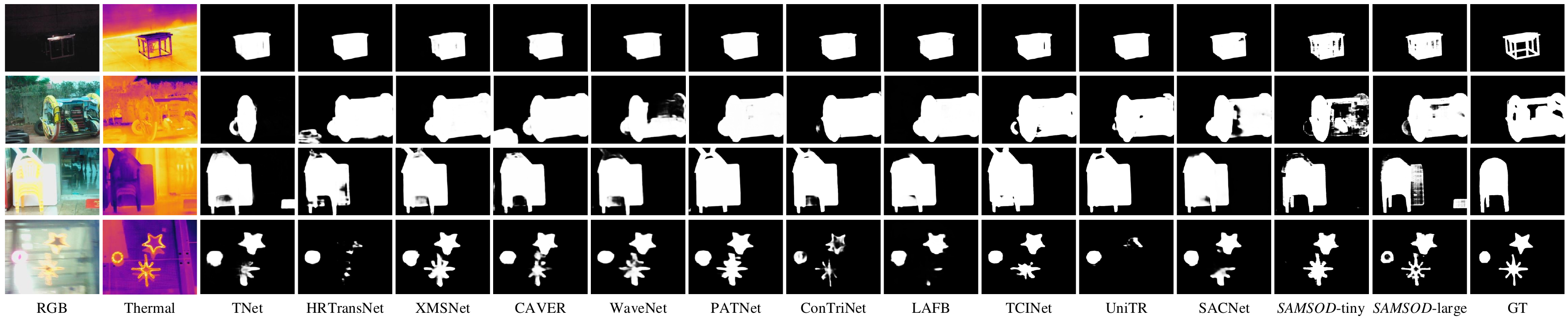}
  \caption{Failure cases involving hollow objects (1$^{st}$ and 2$^{nd}$ rows) and local overexposure scenes (3$^{rd}$ and 4$^{th}$ rows).}
  \label{fig:FailureCases}
\end{figure}

Fig \ref{fig:FailureCases} presents the failure cases. The first two rows show the poor performance when handling hollow objects. Our method primarily focuses on optimizing two modalities while avoiding over-reliance on either one. The segmentation accuracy relies on the fine-tuned SAM, which is based on the Vision Transformer (ViT) architecture. While ViT excels at capturing global context, it is less sensitive to the hollow objects. Additionally, the RGB-T salient object detection datasets are not specifically designed for hollow object segmentation, further limiting the  model's ability in this regard. A potential solution for improving the segmentation ability of hollow objects is to incorporate edge-aware or structure-aware modules to address this limitation. The last two rows display the subpar results in  local overexposure scenes. Local overexposure alters the overall semantics of the RGB image, leading to incorrect recognition of salient objects, such as the white background and red mat in the third row, and the hollow circle in the fourth row. A potential solution is to incorporate diverse exposure augmentation strategies to improve model robustness.

\subsubsection{\textbf{Cost Analysis}}
The last column in Table \ref{tab:RGBTCom} shows the computation cost comparison with the state-of-the-art methods. Specifically, \textit{SAMSOD}-large contains 224.1$M$ parameters, requires 418.0$G$ FLOPs, and runs at 6.4 FPS. It must be acknowledged that while our method achieves good performance, the computational cost is also considerable.
To further meet the requirements of real-time applications, we provide a tiny version which adopts SAM2-tiny in the encoder. It achieves top-three performance while significantly improving real-time efficiency.
Specifically, \textit{SAMSOD}-tiny contains 32.7$M$ parameters, requires 60.4$G$ FLOPs, and runs at 23.1 FPS.
\textit{SAMSOD}-large is better suited for scenarios requiring high accuracy, while \textit{SAMSOD}-tiny offers a more balanced trade-off between performance and computational cost.

\subsection{Ablation Study}
\subsubsection {Optimization Effectiveness Analysis}
Table \ref{tab:opeffectiveness} shows the effectiveness of two optimization solutions. The first row shows the baseline model SAM with vanilla adapters. By introducing the first optimization solution  to address the imbalance convergence issue, the performance in the second row shows a significant improvement. It benefits from unimodal supervision and gradient deconfliction, which enhance the learning of both modalities while also preventing the model from converging to a suboptimal solution, thereby improving robustness.
By introducing the second optimization solution  to address the significant gradient difference between high- and low- activations issue, the performance in the third row show a notable  improvement from the baseline due to decoupled adapters which mitigate the gradient difference between foreground and background, preventing  the background from being ignored during learning.
The comparison between the two groups of optimizations shows that the first group of optimization leads to more significant improvements.
At last, the two groups of optimizations together achieve the best results shown in the fourth row.

\begin{table}[!htp]
\caption{Ablation study on optimizations for addressing the imbalance convergence between  two modalities and significant gradient difference between high- and low- activations.}
\centering
\resizebox{1\linewidth}{!}
{
      \begin{tabular}{ccccccccccccccc}
    \toprule

   {\centering SAM}&{\centering Unimodal Supervsion}&{\centering w/o Vanilla Adapters} & \multicolumn{4}{c}{\centering VT821 } & \multicolumn{4}{c}{\centering VT1000 } & \multicolumn{4}{c}{\centering VT5000 }  \\
   \cmidrule(r){4-7} \cmidrule(r){8-11} \cmidrule{12-15}
    {\centering w/ Vanilla Adapters} &{\centering Gradient Deconfliction}&  {\centering w/ Decoupled Adapters} &$S$$\uparrow$& $F_{\beta}$$\uparrow$& $E_{\xi}$ $\uparrow$& $M\downarrow$
         &$S$$\uparrow$& $F_{\beta}$$\uparrow$& $E_{\xi}$ $\uparrow$& $M\downarrow$
         &$S$$\uparrow$& $F_{\beta}$$\uparrow$& $E_{\xi}$ $\uparrow$& $M\downarrow$
      \\
    \midrule
\checkmark&&
&.915&.899&.946&.025
&.940&.939&.976&.015
&.923&.918&.961&.021
\\

\checkmark&\checkmark&
&.920&.907&.951&.021
&.947&.948&.980&.013
&.932&.929&.967&.018\\
\checkmark&&  \checkmark
&.918&.904&.952&.023
&.941&.942&.981&.014
&.925&.921&.964&.020\\
\checkmark&\checkmark&\checkmark  &\textbf{.924}&\textbf{.916}&\textbf{.956}&\textbf{.021}
      &\textbf{.948}&\textbf{.949}&\textbf{.981}&\textbf{.012}
      &\textbf{.934}&\textbf{.933}&\textbf{.969}&\textbf{.017}\\
    \bottomrule
\end{tabular}
}
\label{tab:opeffectiveness}
\end{table}

\begin{table}[!htp]
\caption{Ablation study on the effectiveness of optimization in addressing the imbalance convergence issue.}
\centering

\resizebox{1\linewidth}{!}
{
   \begin{tabular}{cccccccccccccc}

\hline\toprule
   \multirow{2}{*}{\centering Methods}  &\multicolumn{4}{c}{\centering VT821 } & \multicolumn{4}{c}{\centering VT1000 } & \multicolumn{4}{c}{\centering VT5000 }\\
     \cmidrule(r){2-5} \cmidrule(r){6-9} \cmidrule{10-13}
     & S$\uparrow$
     & F$_\beta$ $\uparrow$ &$E_{\xi}\uparrow$
     & M$\downarrow$ & S$\uparrow$
     & F$_\beta$ $\uparrow$ &$E_{\xi}\uparrow$
     & M$\downarrow$ & S$\uparrow$
     & F$_\beta$$\uparrow$&$E_{\xi}\uparrow$ & M$\downarrow$\\
    \midrule
    Only RGB (SAM+\textbf{VA})
    &.916&.900&.941&.024
    &.945&.947&.978&.016
    &.924&.915&.959&.023\\
    RGB-T (SAM+\textbf{VA})
    &.915&.899&.946&.025
    &.940&.939&.976&.015
    &.923&.918&.961&.021\\
    Add Unimodal Supervision
    &.919&.902&.947&.023
    &.945&.943&.979&.013
    &.930&.925&.966&.019\\
Add Gradient Deconfliction
&.920&.907&.951&\textbf{.021}
&.947&.948&.980&.013
&.932&.929&.967&.018\\
\midrule
    RGB-T (SAM+\textbf{DA})
    &.918&.904&.952&.023
    &.941&.942&.981&.014
    &.925&.921&.964&.020
    \\
    Add Unimodal Supervision
    &.922&.906&.952&.022
    &.946&.948&.979&.014
    &.930&.928&.965&.019\\
    Add Gradient Deconfliction
    &\textbf{.924}&\textbf{.916}&\textbf{.956}&\textbf{.021}
    &\textbf{.948}&\textbf{.949}&\textbf{.981}&\textbf{.012}
    &\textbf{.934}&\textbf{.933}&\textbf{.969}&\textbf{.017}\\
   \bottomrule
    \hline
\end{tabular}
}
\label{tab:op1}
\end{table}
\begin{table}[!htp]
\caption{Generalizability experiment on scribble supervised RGB-T salient object detection dataset.}
\centering

\resizebox{1\linewidth}{!}
{
   \begin{tabular}{cccccccccccccc}

\hline\toprule
   \multirow{2}{*}{\centering Methods}&\multirow{2}{*}{\centering Source} &\multicolumn{4}{c}{\centering VT821 } & \multicolumn{4}{c}{\centering VT1000 } & \multicolumn{4}{c}{\centering VT5000 }\\
     \cmidrule(r){3-6} \cmidrule(r){7-10} \cmidrule{11-14}
     && S$\uparrow$
     & F$_\beta$ $\uparrow$ &$E_{\xi}\uparrow$
     & M$\downarrow$ & S$\uparrow$
     & F$_\beta$ $\uparrow$ &$E_{\xi}\uparrow$
     & M$\downarrow$ & S$\uparrow$
     & F$_\beta$$\uparrow$&$E_{\xi}\uparrow$ & M$\downarrow$\\
    \midrule

    WSVSOD \cite{zhao2021weakly}&CVPR21
    &.822&.762&.875&.052
    &.886&.870&.935&.035
    &.812&.763&.880&.055\\
    DENet \cite{xu2022weakly}&TIP22
    &.813&.732&.839&.054
    &.906&.891&.947&.028
    &.839&.794&.895&.048\\
    RGBTScribble \cite{liu2023RGBTScribble}&ICME23
    &.895&.878&.942&.027
    &.925&.922&.964&.020
    &.877&.859&.933&.033\\

    LGR\cite{wang2024learningLocal}&TCSVT24
    &.893&.870&.941&.034
    &.932&.918&.964&.019
    &.887&.866&.939&.030\\
    Ours (\textit{SAMSOD}-large)&-
    &\textbf{.915}&\textbf{.896}&\textbf{.950}&\textbf{.022}
    &\textbf{.944}&\textbf{.945}&\textbf{.979}&\textbf{.014}
    &\textbf{.916}&\textbf{.908}&\textbf{.958}&\textbf{.023}\\
   \bottomrule
    \hline
\end{tabular}
}
\label{tab:RGBTScribbleCom}
\end{table}

\subsubsection {Optimization Effectiveness about Imbalance Convergence}

\begin{figure}[!htp]
\center
  \includegraphics[width=0.7\linewidth]{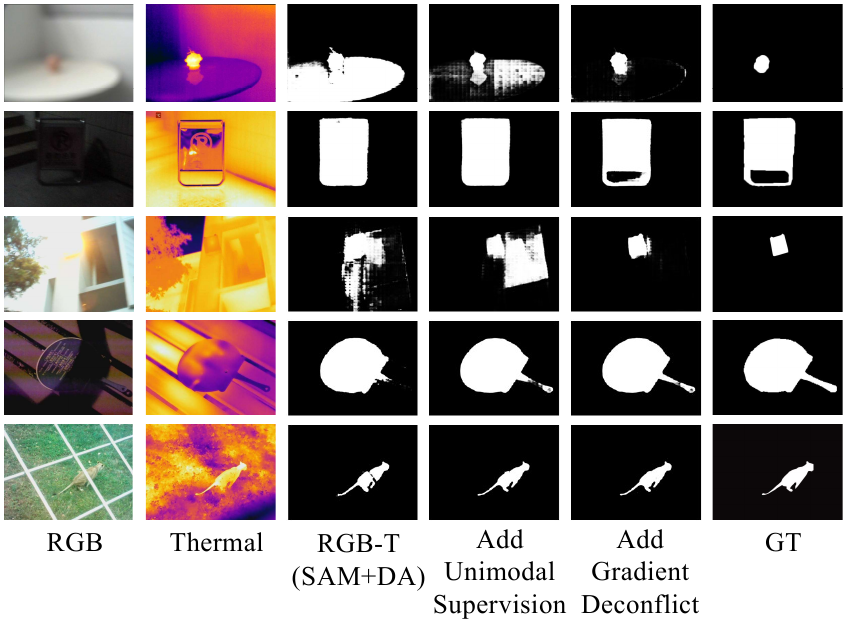}
  \caption{The visualization of the ablation study on unimodal supervision and gradient deconfliction in scenarios with blurring RGB images (1$^{st}$), low light (2$^{nd}$), strong light (3$^{rd}$), shadows (4$^{th}$), and occlusions (5$^{th}$).}
  \label{fig:VisualAblation}
\end{figure}

Table \ref{tab:op1} shows the optimization effectiveness about imbalance convergence issue.
The first row `Only RGB' refers to a single-stream model that uses the RGB modality without incorporating the thermal modality.
The second row `RGB-T (SAM+VA)' refers to a double-stream model that uses two modalities. Both are constructed on SAM model with vanilla adapters.
By the comparison, we find double-steam model is not absolutely better than single-stream model, suggesting that the contribution of the thermal modality has not been fully utilized.
By introducing the unimodal supervision, the thermal modal is well utilized so that the gradient ratio between RGB and thermal encoders is reduced, which can be seen the green curve from Fig \ref{fig:gradRatio}.
Meanwhile, the third row also shows a corresponding improvement, with only 1 out of the 12 metrics decreasing.
Furthermore, gradient deconfliction further improve the performance of salient object detection by reducing potential gradient conflict. The second group of Table \ref{tab:op1} also demonstrates that the unimodal supervision and gradient deconfliction progressively enhance the performance based on decoupled adapters instead of vanilla adapters. Fig \ref{fig:VisualAblation} gives the visualization. In scenarios with blurring RGB images, low light, strong light, shadows, and occlusions, unimodal supervision enhances the learning of the thermal infrared modality, while gradient deconfliction prevents the model from getting stuck in a suboptimal solution, together achieving performance close to the ground truth.

%
%
%

\subsubsection {Optimization Effectiveness about the Mitigation of Significant Gradient Difference}
\begin{figure}[!htp]
\center
  \includegraphics[width=0.7\linewidth]{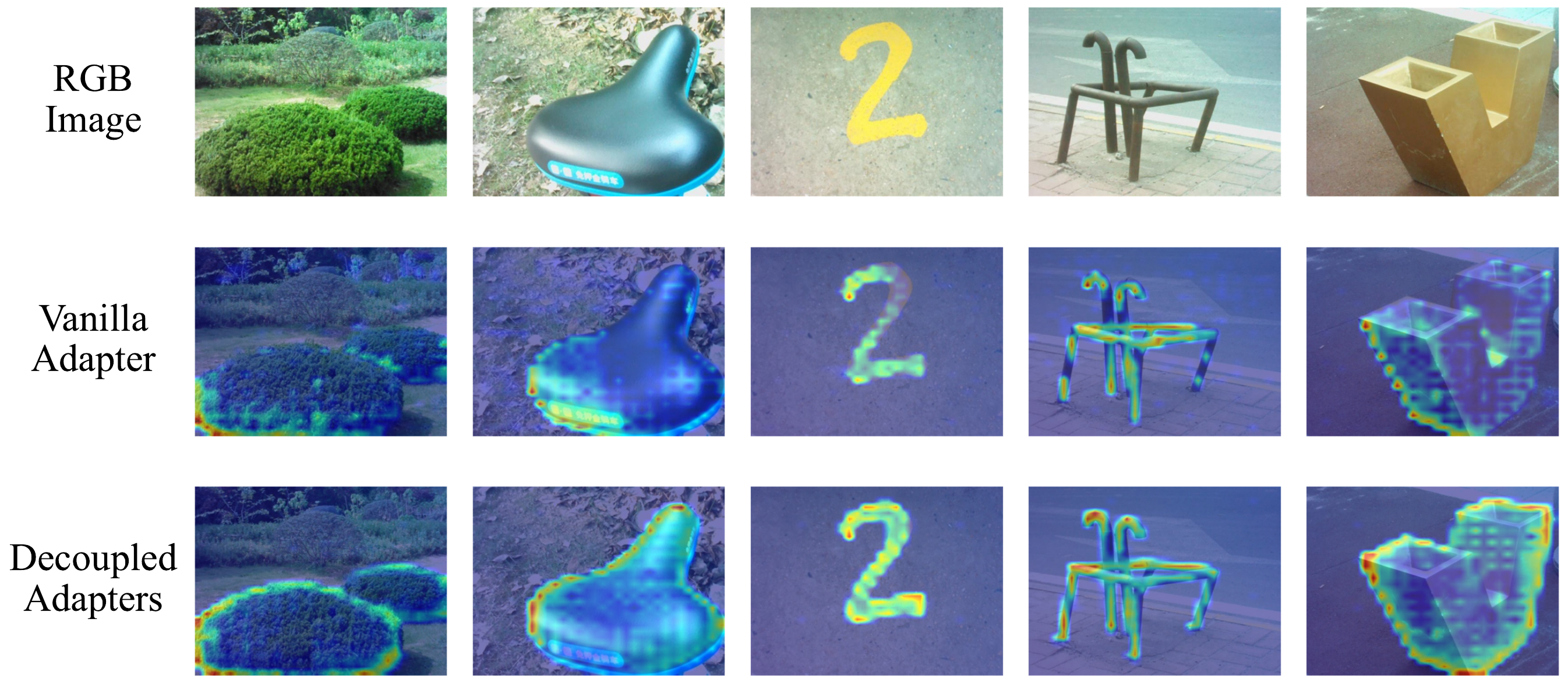}
  \caption{The visual comparison between vanilla adapters (VA) and decoupled adapters (DA).}
  \label{fig:DA}
\end{figure}

Fig \ref{fig:DA} shows the decoupled adapters give a clearer object boundary compared with the vanilla adapter. The quantitative comparison about decoupled adapters and vanilla adapters can be seen in the comparison between the first and third or second and fourth rows of  Table \ref{tab:opeffectiveness}. By zeroing out low- and high- activation neurons in the foreground adapter and background adapter respectively, the gradient interference from high-activation regions on the low-activation pathway is reduced, allowing low-activation features to be adequately updated during training.

%
\subsection{Generalizability Experiments}
Generalizability experiments are conducted on scribble-supervised RGB-T SOD dataset, full-supervised RGB-D SOD dataset, and full-supervised RGB-D  rail surface defect detection. The first is used to test the generalizability from full supervision to weakly supervision, while the last two are used to test the  generalizability from RGB-T task to RGB-D task.

In the first generalizability experiment, RGBT-S dataset \cite{liu2023RGBTScribble} is used to train and test. The comparison methods include WSVSOD \cite{zhao2021weakly}, DENet \cite{xu2022weakly}, RGBTScribble \cite{liu2023RGBTScribble}, and LGR\cite{wang2024learningLocal}. From Table \ref{tab:RGBTScribbleCom}, our method achieves the best performance in scribble supervised task. The excellent results are attributed to the resolution of  imbalance convergence of two modalities and significant gradient difference between high- and low- activations existing in two-stream SOD frameworks, which are not covered by the compared methods.

In the second generalizability experiment, NLPR \cite{peng2014rgbd}, NJU2K \cite{ju2014depth}, STERE\cite{niu2012leveraging}, and SIP \cite{fan2020rethinking} are used to train and test. The concrete partition of training and testing samples follows the common practice.
The comparison methods include  JL-DCF\cite{fu2021siamese}, SwinNet \cite{liu2021swinnet}, 
PCIR-Net \cite{cong2023point}, HRTransNet \cite{tang2022hrtransnet}, HiDANet \cite{wu2023hidanet}, CATNet \cite{sun2023catnet}, TSVT \cite{gao2024tsvt}, 
HFMDNet\cite{luo2024hfmdnet}, MAGNet \cite{zhong2024magnet}, EM-Trans \cite{chen2024trans}, DFormer\cite{yin2023dformer}, TPCL \cite{wu2023transformer}, and DPPNet \cite{yuan2025dppnet}. From Table \ref{tab:RGBDCom}, our method achieves the comparative performance in RGB-D SOD task. The evaluation metric shows a noticeable improvement, especially on STERE and SIP dataset.

\begin{table}[!htp]
\caption{Generalizability experiment on fully supervised RGB-D salient object detection dataset.}
\centering
\resizebox{1\linewidth}{!}
{
   \begin{tabular}{cccccccccccccccccc}
   \hline\toprule
   \multirow{2}{*}{\centering Methods}&\multirow{2}{*}{\centering Source}
   &\multicolumn{4}{c}{\centering NLPR}
   &\multicolumn{4}{c}{\centering NJU2K}
   &\multicolumn{4}{c}{\centering STERE}
   &\multicolumn{4}{c}{\centering SIP}

   \\
   \cmidrule(r){3-6} \cmidrule(r){7-10} \cmidrule{11-14}\cmidrule{15-18}
   && S$\uparrow$& F$_\beta$ $\uparrow$ &$E_{\xi}\uparrow$& M$\downarrow$
   & S$\uparrow$& F$_\beta$ $\uparrow$ &$E_{\xi}\uparrow$& M$\downarrow$
   & S$\uparrow$& F$_\beta$$\uparrow$&$E_{\xi}\uparrow$ & M$\downarrow$
   & S$\uparrow$& F$_\beta$$\uparrow$ &$E_{\xi}\uparrow$& M$\downarrow$

   \\
   \midrule
    JL-DCF\cite{fu2021siamese}&TPAMI21
    &.926&.917&.964&.023
    &.911&.913&.948&.040
    &.911&.907&.949&.039
    &.892&.900&.949&.046

    \\
    SwinNet \cite{liu2021swinnet}&TCSVT22
    &.941&.936&.974&.018
    &.935&.938&.963&.027
    &.919&.918&.956&.033
    &.911&.927&.950&.035

    \\

    PCIR-Net \cite{cong2023point}&ACM MM23
    &.935&.931&.970&.019
    &.927&.931&.958&.029
    &.920&.920&.957&.031
    &.899&.915&.939&.040

    \\
    HRTransNet \cite{tang2022hrtransnet}&TCSVT23
    &.942&.936&.974&.016
    &.933&.939&.963&.026
    &.921&.919&.956&.030
    &.909&.929&.949&.035

    \\
    HiDANet \cite{wu2023hidanet}&TIP23
    &.930&.929&.961&.021
    &.926&.939&.954&.029
    &.911&.921&.946&.035
    &.892&.919&.927&.043

    \\
    CATNet \cite{sun2023catnet}&TMM23
    &.940&.934&.972&.018
    &.932&.937&.960&.026
    &.921&.922&.958&.030
    &.911&.928&.952&.034
\\
   TSVT \cite{gao2024tsvt}&PR24
   &.935&.924&.966&.021
   &.917&.917&.939&.037
   &.912&.915&.946&.035
   &.905&.916&.944&.037\\

   HFMDNet\cite{luo2024hfmdnet}&TIM24
   &.938&.933&.971&.017
   &.937&.944&.966&\textbf{.023}
   &.918&.920&.957&.031
   &.886&.905&.930&.044

   \\
   MAGNet \cite{zhong2024magnet}&KBS24
   &.938&.931&.969&.017
   &.928&.935&.962&.027
   &.922&.920&.956&.029
   &.907&.924&.947&.036\\
   EM-Trans \cite{chen2024trans}&TNNLS24
   &.940&.934&.970&.017
   &.931&.935&.961&.027
   &.925&.926&.958&.028
   &.903&.920&.944&.039\\

   DFormer\cite{yin2023dformer}&ICLR24
    &.942&.936&.973&.016
    &.937&.943&\textbf{.967}&\textbf{.023}
    &.923&.920&.956&.030
    &.915&.930&.953&.032

    \\

   TPCL \cite{wu2023transformer}&TMM24

   &.935&.930&.970&.017
   &.925&.930&.959&.028
   &.916&.917&.956&.031
   &.900&.917&.943&.038\\
   DPPNet \cite{yuan2025dppnet}&TMM25
   &\textbf{.944}&.939&\textbf{.978}&.016
   &.934&.939&.965&.026
   &.922&.921&.956&.032
   &.907&.924&.946&.038\\
   Ours (\textit{SAMSOD}-large)&-
   &.941&\textbf{.942}&.974&\textbf{.015}
   &\textbf{.938}&\textbf{.945}&.966&\textbf{.023}
   &\textbf{.930}&\textbf{.931}&\textbf{.962}&\textbf{.025}
   &\textbf{.920}&\textbf{.937}&\textbf{.955}&\textbf{.029}
  \\

   \bottomrule
    \hline
\end{tabular}
}
\label{tab:RGBDCom}
\end{table}

In the third generalizability experiment, RGB-D  rail surface defect detection on NEU RSDDS-AUG dataset \cite{wang2022collaborative} with 1,500 training pairs and  362 testing ones is adopted.
From Table \ref{tab:surface}, our method achieves excellent performance, matching that of the best method, DSSNet \cite{wang2024depth}. SAM and its optimization for two proposed issues are the main reasons for the performance improvement.
\begin{table}[!htp]
\caption{Generalizability experiment  on NEU RSDDS-AUG dataset \cite{wang2022collaborative} for RGB-D rail surface defect detection.}
\centering
\resizebox{1\linewidth}{!}
{
\begin{tabular}{cccccccc}
  \toprule
    Methods &  Source&$\textit{S}_m\uparrow$&$\textit{maxE}_m\uparrow$&$\textit{maxF}_m\uparrow$&$\textit{meanE}_m\uparrow$&$\textit{meanF}_m\uparrow$&$\textit{MAE}\downarrow$\\
\midrule
CLANet\cite{wang2022collaborative}& TMech22&.834&.921&.877&.912&.863&.069\\
DRERNet\cite{wu2022depth}&SPL22&.844&.933&.891&.929&.878&.059\\
FHENet \cite{zhou2023fhenet}&TIM23&.836&.926&.881&.921&.869&.064\\
CSANet\cite{yang2023csanet}&SPL23&.861&.941&.904&.928&.885&.058\\
MENet\cite{zhou2023modal}&TCSVT23&.856&.939&.899&.931&.884&.057\\
SA2F\cite{huang2023surface}&TII23&.845&.927&.884&.923&.870&.061\\
PENet\cite{wang2023penet}&TIM23&.859&.940&.905&-&-&.054\\
SAINet\cite{yan2024specificity}&OLEN24&.849&.936&.888&.934&.886&.055\\
DSSNet\cite{wang2024depth}&TITS24&.866&.946&.911&\textbf{.942}&\textbf{.897}&.049\\
Ours (\textit{SAMSOD}-large)&-& \textbf{.875} & \textbf{.947} & \textbf{.915} & \textbf{.942} & .896 & \textbf{.048}\\
  \bottomrule
  \end{tabular}}
\label{tab:surface}
\end{table}
\section{Conclusion}
In this paper, we propose a model called \textit{SAMSOD} to achieve RGB-T salient object detection task. \textit{SAMSOD} is a SAM-based method addressing two optimization issues.
To play an equal role of RGB and thermal modalities and avoid imbalance convergence and modality conflict, unimodal supervision and gradient deconfliction are proposed. Furthermore, the vanilla adapters in the SAM are replaced with our proposed decoupled adapters.
The gradient interference from high-activation regions on the low-activation pathway is reduced, allowing the foreground and background to be updated simultaneously and preventing the learning of the background from being weakened.
The final results on RGB-T datasets prove our method's effectiveness. A lightweight version is also provided to meet real-time demand.

\bibliographystyle{IEEEtran}
\bibliography{GradRGBTbibR2}

\vfill

\end{document}